%% file: summarization.tex
\documentclass[11pt]{article}
\usepackage{acl2016}
\usepackage{times}
\usepackage{url}
\usepackage{latexsym}
\usepackage{graphicx} 
\usepackage{color}
\usepackage[T1]{fontenc}

\aclfinalcopy 

\title{Abstractive Text Summarization using Sequence-to-sequence RNNs and Beyond}

 \author{Ramesh Nallapati \\
   IBM Watson \\
   {\tt nallapati@us.ibm.com} \\\And
   Bowen Zhou \\
   IBM Watson \\
   {\tt zhou@us.ibm.com} \\\And
   Cicero  dos Santos \\
   IBM Watson \\
   {\tt cicerons@us.ibm.com} \\\AND
   \c{C}a\u{g}lar G\.{u}l\c{c}ehre \\
   Universit\'e de Montr\'eal \\
   {\tt gulcehrc@iro.umontreal.ca} \\\And
   Bing Xiang \\
   IBM Watson \\
   {\tt bingxia@us.ibm.com} \\
  }

\date{}

\begin{document}
\maketitle
\begin{abstract}
In this work, we model abstractive text summarization using Attentional Encoder-Decoder Recurrent Neural Networks, and show that they achieve state-of-the-art performance on two different corpora. We propose several novel models that address critical problems in summarization that are not adequately modeled by the basic architecture, such as modeling key-words, capturing the hierarchy of sentence-to-word structure, and emitting words that are rare or unseen at training time. Our work shows that many of our proposed models contribute to further improvement in performance. We also propose a new dataset consisting of multi-sentence summaries, and establish performance benchmarks for further research.
\end{abstract}
\input{introduction}

\input{models}

\input{related_work}

\input{experiments}

\input{analysis}

\input{conclusion}

\bibliography{summarization}
\bibliographystyle{acl2016}

\end{document}

%% file: introduction.tex
\section{Introduction}
Abstractive text summarization is the task of generating a headline or a short summary consisting of a few sentences that captures the salient ideas of an article or a passage. We use the adjective `abstractive' to denote a summary that is not a mere selection of a few existing passages or sentences extracted from the source, but a compressed paraphrasing of the main contents of the document, potentially using vocabulary unseen in the source document. 

This task can also be naturally cast as mapping an input sequence of words in a source document to a target sequence of words called summary. In the recent past, deep-learning based models that map an input sequence into another output sequence, called sequence-to-sequence models, have been successful in many problems such as machine translation \cite{nmt}, speech recognition \cite{speech} and video captioning \cite{video_captioning}. In the framework of sequence-to-sequence models, a very relevant model to our task is the attentional Recurrent Neural Network (RNN) encoder-decoder model proposed in 
\newcite{nmt}, which has produced state-of-the-art performance in machine translation (MT), which is also a natural language task.

Despite the similarities, abstractive summarization is a very different problem from MT. Unlike in MT, the target (summary) is typically very short and does not depend very much on the length of the source (document) in summarization. Additionally, a key challenge in summarization is to optimally compress the original document in a {\it lossy manner} such that the key concepts in the original document are preserved, whereas in MT, the translation is expected to be loss-less. In translation, there is a strong notion of almost one-to-one word-level alignment between source and target, but in summarization, it is less obvious. 

We make the following main contributions in this work: (i) We apply the off-the-shelf attentional encoder-decoder RNN that was originally developed for machine translation to summarization, and show that it already outperforms state-of-the-art systems on two different English corpora. (ii) Motivated by concrete problems in summarization that are not sufficiently addressed by the machine translation based model, we propose novel models and show that they provide additional improvement in performance. (iii) We propose a new dataset for the task of abstractive summarization of a document into multiple sentences and establish benchmarks.  

The rest of the paper is organized as follows. In Section \ref{sec:models}, we describe each specific problem in abstractive summarization that we aim to solve, and present a novel model that addresses it. Section \ref{sec:related_work} contextualizes our models with respect to closely related work on the topic of abstractive text summarization. We present the results of our experiments on three different data sets in Section \ref{sec:exp}. We also present some qualitative analysis of the output from our models in Section \ref{sec:analysis} before concluding the paper with remarks on our future direction in Section \ref{sec:conclusion}.



%% file: models.tex
\section{Models}\label{sec:models}
In this section, we first describe the basic encoder-decoder RNN that serves as our baseline and then propose several novel models for summarization, each addressing a specific weakness in the baseline.

\subsection{Encoder-Decoder RNN with Attention and Large Vocabulary Trick}\label{sec:enc_dec}
Our baseline model corresponds to the neural machine translation model used in \newcite{nmt}. The encoder consists of a bidirectional GRU-RNN \cite{gru_rnn}, while the decoder consists of a uni-directional GRU-RNN with the same hidden-state size as that of the encoder, and an attention mechanism over the source-hidden states and a soft-max layer over target vocabulary to generate words. In the interest of space, we refer the reader to the original paper for a detailed treatment of this model.
In addition to the basic model, we also adapted to the summarization problem, the large vocabulary `trick' (LVT) described in \newcite{lvt}. In our approach, the decoder-vocabulary of each mini-batch is restricted to words in the source documents of that batch. In addition, the most frequent words in the target dictionary are added until the vocabulary reaches a fixed size. 
The aim of this technique is to reduce the size of the soft-max layer of the decoder which is the main computational bottleneck. In addition, this technique also speeds up convergence by focusing the modeling effort only on the words that are essential to a given example. This technique is particularly well suited to summarization since a large proportion of the words in the summary come from the source document in any case.


\subsection{Capturing Keywords using Feature-rich Encoder}\label{sec:feats}
\begin{figure}[ht]
    \vspace{-0.3in}
	\centering
  \includegraphics[width=0.5\textwidth]{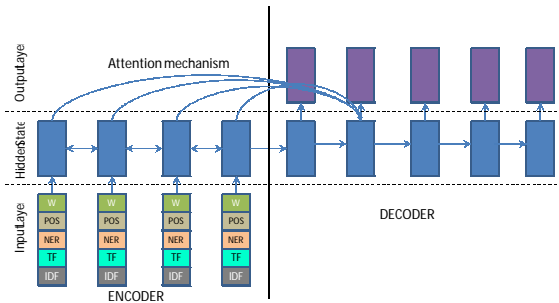}
  \vspace{-0.6in}
	\caption{{\small Feature-rich-encoder: We use one embedding vector each for POS, NER tags and discretized TF and IDF values, which are concatenated together with word-based embeddings as input to the encoder.}}
	\label{fig:feature_rich_encoder}
\end{figure}

In summarization, one of the key challenges is to identify the key concepts and key entities in the document, around which the story revolves. In order to accomplish this goal, we may need to go beyond the word-embeddings-based representation of the input document and capture additional linguistic features such as parts-of-speech tags, named-entity tags, and TF and IDF statistics of the words. We therefore create additional look-up based embedding matrices for the vocabulary of each tag-type, similar to the embeddings for words. For continuous features such as TF and IDF, we convert them into categorical values by discretizing them into a fixed number of bins, and use one-hot representations to indicate the bin number they fall into. This allows us to map them into an embeddings matrix like any other tag-type. Finally, for each word in the source document, we simply look-up its embeddings from all of its associated tags and concatenate them into a single long vector, as shown in Fig. \ref{fig:feature_rich_encoder}.  
On the target side, we continue to use only word-based embeddings as the representation.

\subsection{Modeling Rare/Unseen Words using Switching Generator-Pointer}\label{sec:switch}
\begin{figure}[ht]
    \vspace{-0.3in}
	\centering
  \includegraphics[width=0.5\textwidth]{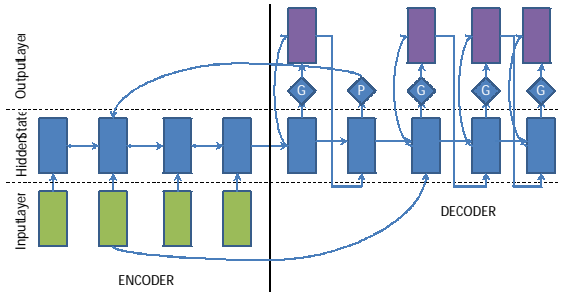}
  \vspace{-0.6in}
	\caption{{\small Switching generator/pointer model: When the switch shows 'G', the traditional generator consisting of the softmax layer is used to produce a word, and when it shows 'P', the pointer network is activated to copy the word from one of the source document positions. When the pointer is activated, the embedding from the source is used as input for the next time-step as shown by the arrow from the encoder to the decoder at the bottom.}}
	\label{fig:switching_generator_pointer}
\end{figure}

Often-times in summarization, the keywords or named-entities in a test document that are central to the summary may actually be unseen or rare with respect to training data. Since the vocabulary of the decoder is fixed at training time, it cannot emit these unseen words. Instead, a most common way of handling these out-of-vocabulary (OOV) words is to emit  an `UNK' token as a placeholder. However this does not result in legible summaries. In summarization, an intuitive way to handle such OOV words is to simply point to their location in the source document instead. We model this notion using our novel switching decoder/pointer architecture which is graphically represented in Figure \ref{fig:switching_generator_pointer}. In this model, the decoder is equipped with a `switch' that decides between using the generator or a pointer at every time-step. If the switch is turned on, the decoder produces a word from its target vocabulary in the normal fashion. However, if the switch is turned off, the decoder instead generates a pointer to one of the word-positions in the source. The word at the pointer-location is then copied into the summary.  The switch is modeled as a sigmoid activation function over a linear layer based on the entire available context at each time-step as shown below.
\begin{eqnarray}
P(s_i=1) &=& \sigma({{\bf v}^s}\cdot({\bf W}^s_h{\bf h}_i + {\bf W}^s_e{\bf E}[o_{i-1}] \nonumber\\
             &+& {\bf W}^s_c{\bf c}_i + {\bf b}^s)),\nonumber
\end{eqnarray}
where $P(s_i=1)$ is the probability of the switch turning on at the $i^{th}$ time-step of the decoder, ${\bf h}_i$ is the hidden state,  ${\bf E}[o_{i-1}]$ is the embedding vector of the emission from the previous time step, ${\bf c_i}$ is the attention-weighted context vector, and ${\bf W}_h^s, {\bf W}_e^s, {\bf W}_c^s, {\bf b}^s$ and ${\bf v}^s$ are the switch parameters. We use attention distribution over word positions in the document as the distribution to sample the pointer from.
\begin{eqnarray}
P^a_i(j)    &\propto&  \exp({{\bf v}^a}\cdot({\bf W}_h^a{\bf h}_{i-1} + {\bf W}^a_e{\bf E}[o_{i-1}] \nonumber\\
            &+& {\bf W}^a_c{\bf h}^d_j + {\bf b}^a)), \nonumber \\
 p_i &=& \arg\max_j (P^a_i(j))~\mbox{for}~j \in \{1,\ldots,N_d\}\nonumber.
\end{eqnarray}
In the above equation, $p_i$ is the pointer value at $i^{th}$ word-position in the summary, sampled from the attention distribution ${\bf P}^a_i$ over the document word-positions $j \in \{1,\ldots,N_d\}$, where $P^a_i(j)$ is the probability of the $i^{th}$ time-step in the decoder pointing to the $j^{th}$ position in the document, and ${\bf h}^d_j$ is the encoder's hidden state at position $j$. 

At training time, we provide the model with explicit pointer information whenever the summary word does not exist in the target vocabulary. When the OOV word in summary occurs in multiple document positions, we break the tie in favor of its first occurrence. At training time, we optimize the conditional log-likelihood shown below, with additional regularization penalties.
\begin{eqnarray}
&& \log P({\bf y}|{\bf x}) = \sum_i (g_i\log \{P(y_i | {\bf y}_{-i}, {\bf x})P(s_i)\} \nonumber\\
 &&+  (1-g_i)\log\{P(p(i)|{\bf y}_{-i},{\bf x})(1-P(s_i))\})\nonumber
\end{eqnarray}
where ${\bf y}$ and ${\bf x}$ are the summary and document words respectively, $g_i$ is an indicator function that is set to 0 whenever the word at position $i$ in the summary is OOV with respect to the decoder vocabulary. At test time, the model decides automatically at each time-step whether to generate or to point, based on the estimated switch probability $P(s_i)$.  We simply use the $\arg\max $ of the posterior probability of generation or pointing to generate the best output at each time step.

The pointer mechanism may be more robust in handling rare words because it uses the encoder's hidden-state representation of rare words to decide which word from the document to point to. Since the hidden state depends on the entire context of the word, the model is able to accurately point to unseen words although they do not appear in the target vocabulary.\footnote{Even when the word does not exist in the source vocabulary, the pointer model may still be able to identify the correct position of the word in the source since it takes into account the contextual representation of the corresponding 'UNK' token encoded by the RNN. Once the position is known, the corresponding token from the source document can be displayed in the summary even when it is not part of the training vocabulary either on the source side or the target side.}  

\subsection{Capturing Hierarchical Document Structure with Hierarchical Attention}\label{sec:hierarchical}
\begin{figure}[ht]
    \vspace{-0.3in}
	\centering
  \includegraphics[width=0.5\textwidth]{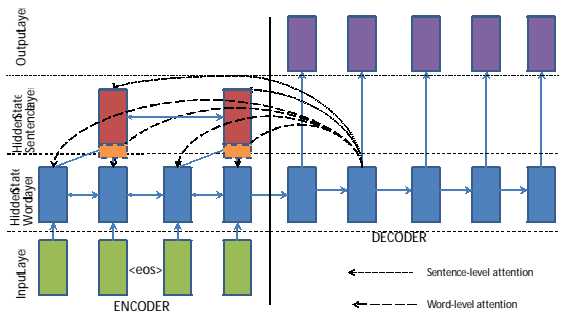}
  	\vspace{-0.6in}
	\caption{{\small Hierarchical encoder with hierarchical attention: the attention weights at the word level, represented by the dashed arrows are re-scaled by the corresponding sentence-level attention weights, represented by the dotted arrows. The dashed boxes at the bottom of the top layer RNN represent sentence-level positional embeddings concatenated to the corresponding hidden states.}}
	\label{fig:hierarchical_attention}
\end{figure}

In datasets where the source document is very long, in addition to identifying the keywords in the document, it is also important to identify the key sentences from which the summary can be drawn. This model aims to capture this notion of two levels of importance using two bi-directional RNNs on the source side, one at the word level and the other at the sentence level. The attention mechanism operates at both levels simultaneously. The word-level attention is further re-weighted by the corresponding sentence-level attention and re-normalized as shown below:
\begin{eqnarray}
P^a(j) &=&  \frac{P^a_w(j)P^a_s(s(j))}{\sum_{k=1}^{N_d} P^a_w(k)P^a_s(s(k))},\nonumber
\end{eqnarray}
where $P^a_w(j)$ is the word-level attention weight at $j^{th}$ position of the source document, and $s(j)$ is the ID of the sentence at $j^{th}$ word position, $P^a_s(l)$ is the sentence-level attention weight for the $l^{th}$ sentence in the source, $N_d$ is the number of words in the source document, and $P^a(j)$ is the re-scaled attention at the  $j^{th}$ word position. The re-scaled attention is then used to compute the attention-weighted context vector that goes as input to the hidden state of the decoder. Further, we also concatenate additional positional embeddings to the hidden state of the sentence-level RNN to model positional importance of sentences in the document. This architecture therefore models key sentences as well as keywords within those sentences jointly. A graphical representation of this model is displayed in Figure \ref{fig:hierarchical_attention}. 

%% file: related_work.tex
\section{Related Work}\label{sec:related_work}

A vast majority of past work in summarization has been extractive, which consists of identifying key sentences or passages in the source document and reproducing them as summary \cite{neto:2002:ATS,extractive,wong:2008:ESU,filippovaA13:EMNLP,colmenares:NAACL2015,graph_based,key_phrases,key_passages}.

Humans on the other hand, tend to paraphrase the original story in their own words. As such, human summaries are abstractive in nature and seldom consist of reproduction of original sentences from the document. The task of abstractive summarization has been standardized using the DUC-2003 and DUC-2004 competitions.\footnote{http://duc.nist.gov/} The data for these tasks consists of news stories from various topics with multiple reference summaries per story generated by humans. 
The best performing system on the DUC-2004 task, called TOPIARY \cite{topiary}, used a combination of linguistically motivated compression techniques, and an unsupervised topic detection algorithm that appends keywords extracted from the article onto the compressed output. Some of the other notable work in the task of abstractive summarization includes using traditional phrase-table based machine translation approaches \cite{mt4summ}, compression using weighted tree-transformation rules \cite{cohn_lapata} and quasi-synchronous grammar approaches \cite{woodsend}.


With the emergence of deep learning as a viable alternative for many NLP tasks \cite{nlp_from_scratch}, researchers have started considering this framework as an  attractive, fully data-driven alternative to abstractive summarization. 
In \newcite{namas}, the authors use convolutional models to encode the source, and a context-sensitive attentional feed-forward neural network to generate the summary, producing state-of-the-art results on Gigaword and DUC datasets. In an extension to this work, \newcite{chopra} used a similar convolutional model for the encoder, but replaced the decoder with an RNN, producing further improvement in performance on both datasets.

In another paper that is closely related to our work, \newcite{hu:2015:EMNLP} introduce a large dataset for Chinese short text summarization. They show promising results on their Chinese dataset using an encoder-decoder RNN, but do not report experiments on English corpora. 

In another very recent work, \newcite{jianpeng} used RNN based encoder-decoder for extractive summarization of documents. This model is not directly comparable to ours since their framework is extractive while ours and that of \cite{namas}, \cite{hu:2015:EMNLP} and \cite{chopra} is abstractive.

Our work starts with the same framework as \cite{hu:2015:EMNLP}, where we use RNNs for both source and target, but we go beyond the standard architecture and propose novel models that address critical problems in summarization. We also note that this work is an extended version of \newcite{nallapati}. In addition to performing more extensive experiments compared to that work, we also propose a novel dataset for document summarization on which we establish benchmark numbers too.

Below, we analyze the similarities and differences of our proposed models with related work on summarization.

\noindent{\bf Feature-rich encoder} (Sec. \ref{sec:feats}): Linguistic features such as POS tags, and named-entities as well as TF and IDF information were used in many extractive approaches to summarization \cite{linguistic_extractive}, but they are novel in the context of deep learning approaches for abstractive summarization, to the best of our knowledge.

\noindent{\bf Switching generator-pointer model} (Sec. \ref{sec:switch}): This model combines extractive and abstractive approaches to summarization in a single end-to-end framework. \newcite{namas} also used a combination of extractive and abstractive approaches, but their extractive model is a separate log-linear classifier with handcrafted features. Pointer networks \cite{pointer_networks} have also been used earlier for the problem of rare words in the context of machine translation \cite{luongACL15}, but the novel addition of switch in our model allows it to strike a balance between when to be faithful to the original source (e.g., for named entities and OOV) and when it is allowed to be creative. We believe such a process arguably mimics how human produces summaries. For a more detailed treatment of this model, and experiments on multiple tasks, please refer to the parallel work published by some of the authors of this work \cite{caglar_acl}.

\noindent{\bf Hierarchical attention model} (Sec. \ref{sec:hierarchical}): Previously proposed hierarchical encoder-decoder models use attention only at sentence-level \cite{hiero_encdec}. The novelty of our approach lies in joint modeling of attention at both sentence and word levels, where the word-level attention is further influenced by sentence-level attention, thus capturing the notion of important sentences and important words within those sentences. Concatenation of positional embeddings with the hidden state at sentence-level is also new.

%% file: experiments.tex
\section{Experiments and Results}\label{sec:exp}

\subsection{Gigaword Corpus}

In this series of experiments\footnote{We used Kyunghyun Cho's code (\url{https://github.com/kyunghyuncho/dl4mt-material}) as the starting point.}, we used the annotated Gigaword corpus as described in \newcite{namas}. We used the scripts made available by the authors of this work\footnote{https://github.com/facebook/NAMAS} to preprocess the data, which resulted in about 3.8M training examples. The script also produces about 400K validation and test examples, but we created a randomly sampled subset of 2000 examples each for validation and testing purposes, on which we report our performance. Further, we also acquired the exact test sample used in \newcite{namas} to make precise comparison of our models with theirs.
We also made small modifications to the script to extract not only the tokenized words, but also system-generated parts-of-speech and named-entity tags. 

\noindent{\bf Training:} For all the models we discuss below, we used 200 dimensional word2vec vectors \cite{word2vec} trained on the same corpus to initialize the model embeddings, but we allowed them to be updated during training. 
The hidden state dimension of the encoder and decoder was fixed at 400 in all our experiments. 
When we used only the first sentence of the document as the source, as done in \newcite{namas}, the encoder vocabulary size was 119,505 and that of the decoder stood at 68,885. We used Adadelta \cite{adadelta} for training, with an initial learning rate of 0.001. We used a batch-size of 50 and randomly shuffled the training data at every epoch, while sorting every 10 batches according to their lengths to speed up training. We did not use any dropout or regularization, but applied gradient clipping. We used early stopping based on the validation set and used the best model on the validation set to report all test performance numbers. 
For all our models, we employ the large-vocabulary trick, where we restrict the decoder vocabulary size to 2,000\footnote{Larger values improved performance only marginally, but at the cost of much slower training.}, because it cuts down the training time per epoch by nearly three times, and helps this and all subsequent models converge in only 50\%-75\% of the epochs needed for the model based on full vocabulary.

\noindent{\bf Decoding}: At decode-time, we used beam search of size 5 to generate the summary, and limited the size of summary to a maximum of 30 words, since this is the maximum size we noticed in the sampled validation set. We found that the average system summary length from all our models (7.8 to 8.3) agrees very closely with that of the ground truth on the validation set (about 8.7 words), without any specific tuning. 

\noindent{\bf Computational costs}: We trained all our models on a single {\it Tesla K40} GPU. Most models took about 10 hours per epoch on an average except the hierarchical attention model, which took 12 hours per epoch. All models typically converged within 15 epochs using our early stopping criterion based on the validation cost. The wall-clock training time until convergence therefore varies between 6-8 days depending on the model. Generating summaries at test time is reasonably fast with a throughput of about 20 summaries per second on a single GPU, using a batch size of 1. 

\noindent{\bf Evaluation metrics}: Similar to \cite{nallapati} and \cite{chopra}, we use the full length F1 variant of Rouge\footnote{\url{http://www.berouge.com/Pages/default.aspx}} to evaluate our system.
Although limited length recall was the preferred metric for most previous work, one of its disadvantages is choosing the length limit which varies from corpus to corpus, making it difficult for researchers to compare performances. Full-length recall, on the other hand, does not impose a length restriction but unfairly favors longer summaries. Full-length F1 solves this problem since it can penalize longer summaries, while not imposing a specific length restriction. 

In addition, we also report the percentage of tokens in the system summary that occur in the source (which we call `src. copy rate' in Table \ref{tab:results}). 

\noindent We describe all our experiments and results on the Gigaword corpus below.


\noindent{\it words-lvt2k-1sent}: This is the baseline attentional encoder-decoder model with the large vocabulary trick. This model is trained only on the first sentence from the source document, as done in \newcite{namas}.

\noindent{\it words-lvt2k-2sent}: This model is identical to the model above except for the fact that it is trained on the first two sentences from the source. On this corpus, adding the additional sentence in the source does seem to aid performance, as shown in Table \ref{tab:results}. We also tried adding more sentences, but the performance dropped, which is probably because the latter sentences in this corpus are not pertinent to the summary.



\noindent{\it words-lvt2k-2sent-hieratt}:  Since we used two sentences from source document, we trained the hierarchical attention model proposed in Sec \ref{sec:hierarchical}. As shown in Table \ref{tab:results}, this model improves performance compared to its flatter counterpart by learning the relative importance of the first two sentences automatically. 


\noindent{\it feats-lvt2k-2sent}:  Here, we still train on the first two sentences, but we exploit the parts-of-speech and named-entity tags in the annotated gigaword corpus as well as TF, IDF values, to augment the input embeddings on the source side as described in Sec \ref{sec:feats}.
In total, our embedding vector grew from the original 100 to 155, and produced incremental gains compared to its counterpart {\it words-lvt2k-2sent} as shown in Table \ref{tab:results}, demonstrating the utility of syntax based features in this task. 


\noindent{\it feats-lvt2k-2sent-ptr}: This is the switching generator/pointer model described in Sec. \ref{sec:switch}, but in addition, we also use feature-rich embeddings on the document side as in the above model.
Our experiments indicate that the new model is able to achieve the best performance on our test set by all three Rouge variants as shown in Table \ref{tab:results}.


\noindent{\bf Comparison with state-of-the-art:} 
We compared the performance of our model {\it words-lvt2k-1sent} with state-of-the-art models on the sample created by \newcite{namas}, as displayed in the bottom part of Table \ref{tab:results}. We also trained another system which we call {\it words-lvt5k-1sent} which has a larger LVT vocabulary size of 5k, but also has much larger source and target vocabularies of 400K and 200K respectively.

The reason we did not evaluate our best validation models here is that this test set consisted of only 1 sentence from the source document, and did not include NLP annotations, which are needed in our best models. The table shows that, despite this fact, our model outperforms the ABS+ model of \newcite{namas} with statistical significance. In addition, our models exhibit better abstractive ability as shown by the {\it src. copy rate} metric in the last column of the table. Further, our larger model {\it words-lvt5k-1sent} outperforms the state-of-the-art model of \cite{chopra} with statistically significant improvement on Rouge-1. 

We believe the bidirectional RNN we used to model the source captures richer contextual information of every word than the bag-of-embeddings representation used by \newcite{namas} and \newcite{chopra} in their convolutional attentional encoders, which might explain our superior performance. 
Further, explicit modeling of important information such as multiple source sentences, word-level linguistic features, using the switch mechanism to point to source words when needed, and hierarchical attention, solve specific problems in summarization, each boosting performance incrementally. 



\begin{table*}
\begin{center}
{\small
\begin{tabular}{|r|l|r|r|r|r|}
\hline
{\bf \#} &  {\bf Model name} & {\bf Rouge-1} & {\bf Rouge-2} & {\bf Rouge-L} & {\bf Src. copy rate (\%)}\\
\hline

\hline


\hline
\hline
 \multicolumn{6}{|c|}{Full length F1 on our internal test set}  \\
\hline

1 & words-lvt2k-1sent      &    34.97              &    17.17                 &   32.70           &   75.85  \\
2 & words-lvt2k-2sent      &     35.73             &    17.38                 &   33.25           &    79.54 \\
3 & words-lvt2k-2sent-hieratt &   36.05          &    18.17                 &   33.52            &    78.52   \\
4 & feats-lvt2k-2sent             &   35.90           &    17.57                &   33.38         &      78.92          \\
5 & feats-lvt2k-2sent-ptr   &   *{\bf 36.40}            &    {\bf 17.77}                 &        *{\bf 33.71}   & 78.70  \\
\hline
\multicolumn{6}{|c|}{Full length F1 on the test set used by \cite{namas}}  \\
\hline
6 & ABS+ \cite{namas}  &    29.78      &      11.89             &     26.97        &   91.50       \\
7 & words-lvt2k-1sent & 32.67 &  15.59   & 30.64   &  74.57     \\
8 & RAS-Elman \cite{chopra} & 33.78 &  15.97  & 31.15 & \\
9 & words-lvt5k-1sent & *{\bf 35.30} & {\bf 16.64} & *{\bf 32.62} & \\
\hline
\end{tabular}
}
\end{center}
\vspace{-0.1in}
\caption{{\small Performance comparison of various models. '*' indicates statistical significance of the corresponding model with respect to the baseline model on its dataset as given by the 95\% confidence interval in the official Rouge script. We report statistical significance only for the best performing models. 'src. copy rate' for the reference data on our validation sample is 45\%. Please refer to Section \ref{sec:exp} for explanation of notation.}}
\label{tab:results}
\vspace{-0.2in}
\end{table*}

\subsection{DUC Corpus}

The DUC corpus\footnote{{\it http://duc.nist.gov/duc2004/tasks.html}} comes in two parts: the 2003 corpus consisting of 624 document, summary pairs and the 2004 corpus consisting of 500 pairs. Since these corpora are too small to train large neural networks on, \newcite{namas} trained their models on the Gigaword corpus, but combined it with an additional log-linear extractive summarization model with handcrafted features, that is trained on the DUC 2003 corpus. They call the original neural attention model the ABS model, and the combined model ABS+. \newcite{chopra} also report the performance of their RAS-Elman model on this corpus and is the current state-of-the-art since it outperforms all previously published baselines including non-neural network based extractive and abstractive systems, as measured by the official DUC metric of recall at 75 bytes. In these experiments, we use the same metric to evaluate our models too, but we omit reporting numbers from other systems in the interest of space.

In our work, we simply run the models trained on Gigaword corpus as they are, without tuning them on the DUC validation set. The only change we made to the decoder is to suppress the model from emitting the end-of-summary tag, and force it to emit exactly 30 words for every summary, since the official evaluation on this corpus is based on limited-length Rouge recall. On this corpus too, since we have only a single sentence from source and no NLP annotations, we ran just the models {\it words-lvt2k-1sent} and {\it words-lvt5k-1sent}.

The performance of this model on the test set is compared with ABS and ABS+ models, RAS-Elman from \cite{chopra}, as well as TOPIARY, the top performing system on DUC-2004 in Table \ref{tab:duc}. We note our best model {\it words-lvt5k-1sent} outperforms RAS-Elman on two of the three variants of Rouge, while being competitive on Rouge-1.

\begin{table}[h]
\centering
{\small
\begin{tabular}{|l|r|r|r|}
\hline
Model & Rouge-1 & Rouge-2 & Rouge-L \\
\hline
TOPIARY & 25.12 & 6.46 & 20.12 \\
ABS &   26.55 & 7.06 & 22.05 \\
ABS+ &   28.18 & 8.49 & 23.81 \\
RAS-Elman & {\bf 28.97} & 8.26 & 24.06 \\
words-lvt2k-1sent &  28.35 & {\bf 9.46} & 24.59 \\
words-lvt5k-1sent & 28.61 & 9.42 &  {\bf 25.24} \\
\hline
\end{tabular}
}
\caption{{\small Evaluation of our models using the limited-length Rouge Recall  at 75 bytes on DUC validation and test sets. Our best model, although trained exclusively on the Gigaword corpus, consistently outperforms the ABS+ model which is tuned on the DUC-2003 validation corpus in addition to being trained on the Gigaword corpus.}}
\label{tab:duc}
\end{table}

\subsection{CNN/Daily Mail Corpus}
The existing abstractive text summarization corpora including Gigaword and DUC consist of only one sentence in each summary. In this section, we present a new corpus that comprises multi-sentence summaries. To produce this corpus, we modify an existing corpus that has been used for the task of passage-based question answering \cite{reading_comprehension}. In this work, the authors used the human generated abstractive summary bullets from new-stories in {\it CNN} and {\it Daily Mail} websites as questions (with one of the entities hidden), and stories as the corresponding passages from which the system is expected to answer the fill-in-the-blank question. The authors released the scripts that crawl,  extract and generate pairs of passages and questions from these websites. With a simple modification of the script, we restored all the summary bullets of each story in the original order to obtain a multi-sentence summary, where each bullet is treated as a sentence. In all, this corpus has 286,817 training pairs, 13,368 validation pairs and 11,487 test pairs, as defined by their scripts. The source documents in the training set have 766 words spanning 29.74 sentences on an average while the summaries consist of 53 words and 3.72 sentences. The unique characteristics of this dataset such as long documents, and ordered multi-sentence summaries present interesting challenges, and we hope will attract future researchers to build and test novel models on it.

The dataset is released in two versions: one consisting of actual entity names, and the other, in which entity occurrences are replaced with document-specific integer-ids beginning from 0. Since the vocabulary size is smaller in the anonymized version, we used it in all our experiments below. We limited the source vocabulary size to 150K, and the target vocabulary to 60K, the source and target lengths to at most 800 and 100 words respectively. We used 100-dimensional word2vec embeddings trained on this dataset as input, and we fixed the model hidden state size at 200. We also created explicit pointers in the training data by matching only the anonymized entity-ids between source and target on similar lines as we did for the OOV words in Gigaword corpus. 

\begin{table}[]
\centering
{\small
\begin{tabular}{|l|r|r|r|}
\hline
Model & Rouge-1 & Rouge-2 & Rouge-L \\
\hline
words-lvt2k & 32.49 & 11.84 & 29.47 \\
words-lvt2k-hieratt & 32.75 & 12.21 & 29.01 \\
words-lvt2k-temp-att & *{\bf 35.46} & *{\bf 13.30} & *{\bf 32.65} \\
\hline
\end{tabular}
}
\caption{{\small Performance of various models on CNN/Daily Mail test set using full-length Rouge-F1 metric. Bold faced numbers indicate best performing system.}}
\label{tab:cnn}
\vspace{-0.2in}
\end{table}

\begin{table*}[htpb]
\centering
\begin{tabular}{|p{15cm}|}
\hline
{\bf Source Document}\\
\hline
( @entity0 ) wanted : film director , must be eager to shoot footage of golden lassos and invisible jets . <eos> @entity0 confirms that @entity5 is leaving the upcoming " @entity9 " movie ( the hollywood reporter first broke the story ) . <eos> @entity5 was announced as director of the movie in november . <eos> @entity0 obtained a statement from @entity13 that says , " given creative differences , @entity13 and @entity5 have decided not to move forward with plans to develop and direct ' @entity9 ' together . <eos> " ( @entity0 and @entity13 are both owned by @entity16 . <eos> ) the movie , starring @entity18 in the title role of the @entity21 princess , is still set for release on june 00 , 0000 . <eos> it 's the first theatrical movie centering around the most popular female superhero . <eos> @entity18 will appear beforehand in " @entity25 v. @entity26 : @entity27 , " due out march 00 , 0000 . <eos> in the meantime , @entity13 will need to find someone new for the director 's chair . <eos> \\
\hline
{\bf Ground truth Summary}\\
\hline
@entity5 is no longer set to direct the first " @entity9 " theatrical movie <eos> @entity5 left the project over " creative differences " <eos> movie is currently set for 0000 \\
\hline
{\bf words-lvt2k}\\
\hline
@entity0 confirms that @entity5 is leaving the upcoming " @entity9 " movie <eos> @entity13 and @entity5 have decided not to move forward with plans to develop <eos> @entity0 confirms that @entity5 is leaving the upcoming " @entity9 " movie \\
\hline
{\bf words-lvt2k-hieratt}\\
\hline
@entity5 is leaving the upcoming " @entity9 " movie <eos> the movie is still set for release on june 00 , 0000 <eos> @entity5 is still set for release on june 00 , 0000 \\
\hline
{\bf words-lvt2k-temp-att}\\
\hline
@entity0 confirms that @entity5 is leaving the upcoming " @entity9 " movie <eos> the movie is the first film to around the most popular female actor <eos> @entity18 will appear in " @entity25 , " due out march 00 , 0000 \\
\hline
\end{tabular}
\caption{Comparison of gold truth summary with summaries from various systems. Named entities and numbers are anonymized by the preprocessing script. The "<eos>" tags represent the boundary between two highlights. The temporal attention model ({\it words-lvt2k-temp-att}) solves the problem of repetitions in summary as exhibited by the models {\it words-lvt2k} and {\it words-lvt2k-hieratt} by encouraging the attention model to focus on the uncovered portions of the document.}
\label{tab:cnn_example}
\end{table*}

\noindent{\bf Computational costs:} We used a single Tesla K-40 GPU to train our models on this dataset as well. While the flat models ({\it words-lvt2k} and {\it words-lvt2k-ptr}) took under 5 hours per epoch, the hierarchical attention model was very expensive, consuming nearly 12.5 hours per epoch. 
Convergence of all models is also slower on this dataset compared to Gigaword, taking nearly 35 epochs for all models. Thus, the wall-clock time for training until convergence is about 7 days for the flat models, but nearly 18 days for the hierarchical attention model. Decoding is also slower as well, with a throughput of 2 examples per second for flat models and 1.5 examples per second for the hierarchical attention model, when run on a single GPU with a batch size of 1.

\noindent{\bf Evaluation:} We evaluated our models using the full-length Rouge F1 metric that we employed for the Gigaword corpus, but with one notable difference: in both system and gold summaries, we considered each highlight to be a separate sentence.\footnote{On this dataset, we used the {\it pyrouge} script (\url{https://pypi.python.org/pypi/pyrouge/0.1.0}) that allows evaluation of each sentence as a separate unit. Additional pre-processing involves assigning each highlight to its own "<a>" tag in the system and gold xml files that go as input to the Rouge evaluation script. Similar evaluation was also done by \cite{jianpeng}.}

\noindent{\bf Results:} Results from the basic attention encoder-decoder as well as the hierarchical attention model are displayed in Table \ref{tab:cnn}. Although this dataset is smaller and more complex than the Gigaword corpus, it is interesting to note that the Rouge numbers are in the same range. However, the hierarchical attention model described in Sec. \ref{sec:hierarchical} outperforms  the baseline attentional decoder only marginally. 

Upon visual inspection of the system output, we noticed that on this dataset, both these models  produced summaries that contain repetitive phrases or even repetitive sentences at times. Since the summaries in this dataset involve multiple sentences, it is likely that the decoder `forgets' what part of the document was used in producing earlier highlights. To overcome this problem, we used the {\it Temporal Attention} model of \newcite{baskaran} that keeps track of past attentional weights of the decoder and expliticly discourages it from attending to the same parts of the document in future time steps. The model works as shown by the following simple equations:
\begin{equation}
{\bf \beta}_t = \sum_{k=1}^{t-1}{\bf \alpha}_k';~~ {\bf \alpha}_t \propto \frac{{\bf \alpha}'_t}{{\bf \beta}_t} \nonumber
\end{equation}
where ${\bf \alpha}'_t$ is the unnormalized attention-weights vector at the $t^{th}$ time-step of the decoder. In other words, the temporal attention model down-weights the attention weights at the current time step if the past attention weights are high on the same part of the document. 

Using this strategy, the temporal attention model improves performance significantly over both the baseline model as well as the hierarchical attention model. We have also noticed that there are fewer repetitions of summay highlights produced by this model as shown in the example in Table \ref{tab:cnn_example}.

These results, although preliminary, should serve as a good baseline for future researchers to compare their models against. 


%% file: analysis.tex
\section{Qualitative Analysis}\label{sec:analysis}

Table \ref{fig:example_outputs} presents a few high quality and poor quality output on the validation set from {\it feats-lvt2k-2sent}, one of our best performing models. Even when the model differs from the target summary, its summaries tend to be very meaningful and relevant, a phenomenon not captured by word/phrase matching evaluation metrics such as Rouge. On the other hand, the model sometimes `misinterprets' the semantics of the text and generates a summary with a comical interpretation as shown in the poor quality examples in the table. Clearly, capturing the `meaning' of complex sentences remains a weakness of these models.

\begin{table}
\begin{center}
{\small
\begin{tabular}{|p{7.5cm}|}
\hline
{\bf Good quality summary output} \\
\hline
{\bf S}: a man charged with the murder last year of a british backpacker confessed to the slaying on the night he was charged with her killing , according to police evidence presented at a court hearing tuesday .  ian douglas previte , \#\# , is charged with murdering caroline stuttle , \#\# , of yorkshire , england\\ 
{\bf T}: man charged with british backpacker 's death confessed to crime police officer claims \\
{\bf O}: man charged with murdering british backpacker confessed to murder \\
 \hline
{\bf S}: following are the leading scorers in the english premier league after saturday 's matches : \#\# - alan shearer -lrb- newcastle united -rrb- , james beattie .\\
{\bf T}: leading scorers in english premier league \\
{\bf O}: english premier league leading scorers \\
\hline
{\bf S}: volume of transactions at the nigerian stock exchange has continued its decline since last week , a nse official said thursday .  the latest statistics showed that a total of \#\#.\#\#\# million shares valued at \#\#\#.\#\#\# million naira -lrb- about \#.\#\#\# million us dollars -rrb- were traded on wednesday in \#,\#\#\# deals .\\
{\bf T}: transactions dip at nigerian stock exchange \\
{\bf O}: transactions at nigerian stock exchange down \\
\hline
{\bf Poor quality summary output} \\
\hline
{\bf S}: broccoli and broccoli sprouts contain a chemical that kills the bacteria responsible for most stomach cancer , say researchers , confirming the dietary advice that moms have been handing out for years . in laboratory tests the chemical , $<$unk$>$ , killed helicobacter pylori , a bacteria that causes stomach ulcers and often fatal stomach cancers . \\
{\bf T}:  for release at \#\#\#\# $<$unk$>$ mom was right broccoli is good for you say cancer researchers \\
{\bf O}: broccoli sprouts contain deadly bacteria \\
\hline
{\bf S}: norway delivered a diplomatic protest to russia on monday after three norwegian fisheries research expeditions were barred from russian waters . the norwegian research ships were to continue an annual program of charting fish resources shared by the two countries in the barents sea region .\\
{\bf T}: norway protests russia barring fisheries research ships \\
{\bf O}: norway grants diplomatic protest to russia\\
\hline
{\bf S}: j.p. morgan chase 's ability to recover from a slew of recent losses rests largely in the hands of two men , who are both looking to restore tarnished reputations and may be considered for the top job someday .  geoffrey <unk> , now the co-head of j.p. morgan 's investment bank , left goldman , sachs \& co. more than a decade ago after executives say he lost out in a bid to lead that firm .\\
{\bf T}: \# executives to lead j.p. morgan chase on road to recovery \\
{\bf O}: j.p. morgan chase may be considered for top job \\
\hline
\end{tabular}
}
\end{center}
\caption{\label{fig:example_outputs} {\small Examples of generated summaries from our best model on the validation set of Gigaword corpus. {\bf S}: source document, {\bf T}: target summary, {\bf O}: system output. Although we displayed equal number of good quality and poor quality summaries in the table, the good ones are far more prevalent than the poor ones.}}
\vspace{-2mm}
\end{table}

\begin{figure} 
	\centering
  \includegraphics[width=.5\textwidth]{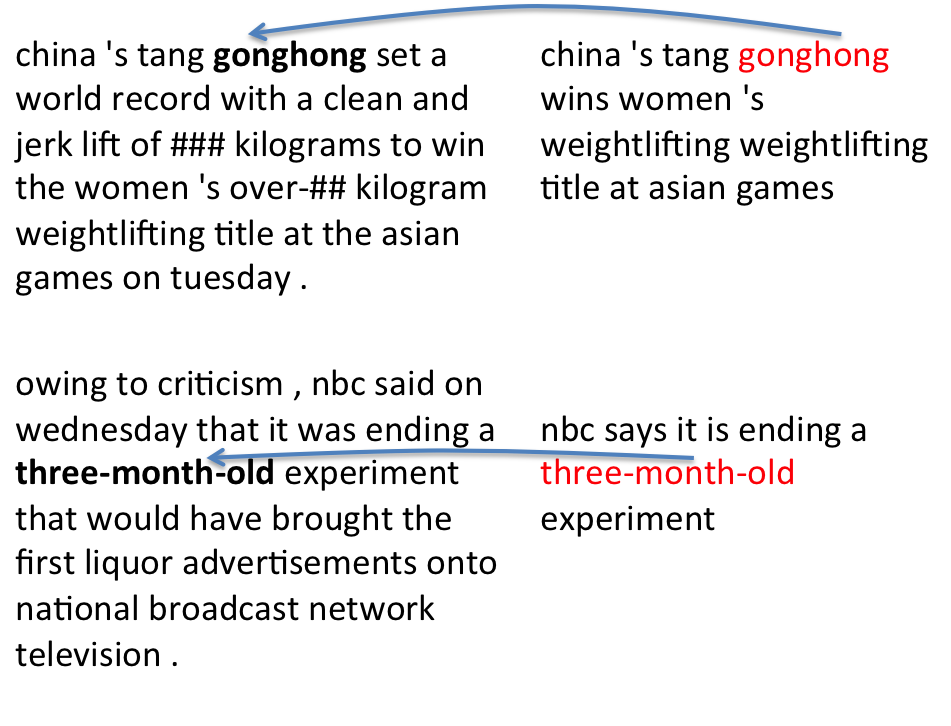}
 \vspace{-0.4in}
\caption{{\small Sample output from switching generator/pointer networks. An arrow indicates that a pointer to the source position was used to generate the corresponding summary word.}}
\label{fig:example_pointers}	
\end{figure}

Our next example output, presented in Figure \ref{fig:example_pointers}, displays the sample output from the switching generator/pointer model on the Gigaword corpus. It is apparent from the examples that the model learns to use pointers very accurately not only for named entities, but also for multi-word phrases. Despite its accuracy, the performance improvement of the overall model is not significant. We believe the impact of this model may be more pronounced in other settings with a heavier tail distribution of rare words. We intend to carry out more experiments with this model in the future.

On CNN/Daily Mail data, although our models are able to produce good quality multi-sentence summaries, we notice that the same sentence or phrase often gets repeated in the summary. We believe models that incorporate intra-attention such as \newcite{lstmn} can fix this problem by encouraging the model to `remember' the words it has already produced in the past.

%% file: conclusion.tex
\section{Conclusion}\label{sec:conclusion}
In this work, we apply the attentional encoder-decoder for the task of abstractive summarization with very promising results, outperforming state-of-the-art results significantly on two different datasets. Each of our proposed novel models addresses a specific problem in abstractive summarization, yielding further improvement in performance. We also propose a new dataset for multi-sentence summarization and establish benchmark numbers on it. As part of our future work, we plan to focus our efforts on this data and build more robust models for summaries consisting of multiple sentences.


%% file: summarization.bbl
\begin{thebibliography}{}

\bibitem[\protect\citename{Bahdanau \bgroup et al.\egroup }2014]{nmt}
Dzmitry Bahdanau, Kyunghyun Cho, and Yoshua Bengio.
\newblock 2014.
\newblock Neural machine translation by jointly learning to align and
  translate.
\newblock {\em CoRR}, abs/1409.0473.

\bibitem[\protect\citename{Bahdanau \bgroup et al.\egroup }2015]{speech}
Dzmitry Bahdanau, Jan Chorowski, Dmitriy Serdyuk, Philemon Brakel, and Yoshua
  Bengio.
\newblock 2015.
\newblock End-to-end attention-based large vocabulary speech recognition.
\newblock {\em CoRR}, abs/1508.04395.

\bibitem[\protect\citename{Banko \bgroup et al.\egroup }2000]{mt4summ}
Michele Banko, Vibhu~O. Mittal, and Michael~J Witbrock.
\newblock 2000.
\newblock Headline generation based on statistical translation.
\newblock {\em In Proceedings of the 38th Annual Meeting on Association for
  Computational Linguistics}, 22:318--325.

\bibitem[\protect\citename{Cheng and Lapata}2016]{jianpeng}
Jianpeng Cheng and Mirella Lapata.
\newblock 2016.
\newblock Neural summarization by extracting sentences and words.
\newblock In {\em Proceedings of the 54th Annual Meeting of the Association for
  Computational Linguistics}.

\bibitem[\protect\citename{Cheng \bgroup et al.\egroup }2016]{lstmn}
Jianpeng Cheng, Li~Dong, and Mirella Lapata.
\newblock 2016.
\newblock Long short-term memory-networks for machine reading.
\newblock {\em CoRR}, abs/1601.06733.

\bibitem[\protect\citename{Chopra \bgroup et al.\egroup }2016]{chopra}
Sumit Chopra, Michael Auli, and Alexander~M. Rush.
\newblock 2016.
\newblock Abstractive sentence summarization with attentive recurrent neural
  networks.
\newblock In {\em HLT-NAACL}.

\bibitem[\protect\citename{Chung \bgroup et al.\egroup }2014]{gru_rnn}
Junyoung Chung, {\c{C}}aglar G{\"{u}}l{\c{c}}ehre, KyungHyun Cho, and Yoshua
  Bengio.
\newblock 2014.
\newblock Empirical evaluation of gated recurrent neural networks on sequence
  modeling.
\newblock {\em CoRR}, abs/1412.3555.

\bibitem[\protect\citename{Cohn and Lapata}2008]{cohn_lapata}
Trevor Cohn and Mirella Lapata.
\newblock 2008.
\newblock Sentence compression beyond word deletion.
\newblock In {\em Proceedings of the 22Nd International Conference on
  Computational Linguistics - Volume 1}, pages 137--144.

\bibitem[\protect\citename{Collobert \bgroup et al.\egroup
  }2011]{nlp_from_scratch}
Ronan Collobert, Jason Weston, L{\'{e}}on Bottou, Michael Karlen, Koray
  Kavukcuoglu, and Pavel~P. Kuksa.
\newblock 2011.
\newblock Natural language processing (almost) from scratch.
\newblock {\em CoRR}, abs/1103.0398.

\bibitem[\protect\citename{Colmenares \bgroup et al.\egroup
  }2015]{colmenares:NAACL2015}
Carlos~A. Colmenares, Marina Litvak, Amin Mantrach, and Fabrizio Silvestri.
\newblock 2015.
\newblock Heads: Headline generation as sequence prediction using an abstract
  feature-rich space.
\newblock In {\em Proceedings of the 2015 Conference of the North American
  Chapter of the Association for Computational Linguistics: Human Language
  Technologies}, pages 133--142.

\bibitem[\protect\citename{Erkan and Radev}2004]{extractive}
G.~Erkan and D.~R. Radev.
\newblock 2004.
\newblock Lexrank: Graph-based lexical centrality as salience in text
  summarization.
\newblock {\em Journal of Artificial Intelligence Research}, 22:457--479.

\bibitem[\protect\citename{Filippova and Altun}2013]{filippovaA13:EMNLP}
Katja Filippova and Yasemin Altun.
\newblock 2013.
\newblock Overcoming the lack of parallel data in sentence compression.
\newblock In {\em Proceedings of the 2013 Conference on Empirical Methods in
  Natural Language Processing}, pages 1481--1491.

\bibitem[\protect\citename{Gulcehre \bgroup et al.\egroup }2016]{caglar_acl}
Caglar Gulcehre, Sungjin Ahn, Ramesh Nallapati, Bowen Zhou, and Yoshua Bengio.
\newblock 2016.
\newblock Pointing the unknown words.
\newblock In {\em Proceedings of the 54th Annual Meeting of the Association for
  Computational Linguistics}.

\bibitem[\protect\citename{Hermann \bgroup et al.\egroup
  }2015]{reading_comprehension}
Karl~Moritz Hermann, Tom{\'{a}}s Kocisk{\'{y}}, Edward Grefenstette, Lasse
  Espeholt, Will Kay, Mustafa Suleyman, and Phil Blunsom.
\newblock 2015.
\newblock Teaching machines to read and comprehend.
\newblock {\em CoRR}, abs/1506.03340.

\bibitem[\protect\citename{Hu \bgroup et al.\egroup }2015]{hu:2015:EMNLP}
Baotian Hu, Qingcai Chen, and Fangze Zhu.
\newblock 2015.
\newblock Lcsts: A large scale chinese short text summarization dataset.
\newblock In {\em Proceedings of the 2015 Conference on Empirical Methods in
  Natural Language Processing}, pages 1967--1972, Lisbon, Portugal, September.
  Association for Computational Linguistics.

\bibitem[\protect\citename{Jean \bgroup et al.\egroup }2014]{lvt}
S{\'{e}}bastien Jean, Kyunghyun Cho, Roland Memisevic, and Yoshua Bengio.
\newblock 2014.
\newblock On using very large target vocabulary for neural machine translation.
\newblock {\em CoRR}, abs/1412.2007.

\bibitem[\protect\citename{K.~Riedhammer and Hakkani-Tur}2010]{key_phrases}
B.~Favre K.~Riedhammer and D.~Hakkani-Tur.
\newblock 2010.
\newblock Long story short – global unsupervised models for keyphrase based
  meeting summarization.
\newblock In {\em Speech Communication}, pages 801--815.

\bibitem[\protect\citename{Li \bgroup et al.\egroup }2015]{hiero_encdec}
Jiwei Li, Minh{-}Thang Luong, and Dan Jurafsky.
\newblock 2015.
\newblock A hierarchical neural autoencoder for paragraphs and documents.
\newblock {\em CoRR}, abs/1506.01057.

\bibitem[\protect\citename{Litvak and Last}2008]{graph_based}
M.~Litvak and M.~Last.
\newblock 2008.
\newblock Graph-based keyword extraction for single-document summarization.
\newblock In {\em Coling 2008}, pages 17--24.

\bibitem[\protect\citename{Luong \bgroup et al.\egroup }2015]{luongACL15}
Thang Luong, Ilya Sutskever, Quoc~V. Le, Oriol Vinyals, and Wojciech Zaremba.
\newblock 2015.
\newblock Addressing the rare word problem in neural machine translation.
\newblock In {\em Proceedings of the 53rd Annual Meeting of the Association for
  Computational Linguistics and the 7th International Joint Conference on
  Natural Language Processing of the Asian Federation of Natural Language
  Processing}, pages 11--19.

\bibitem[\protect\citename{Mikolov \bgroup et al.\egroup }2013]{word2vec}
Tomas Mikolov, Ilya Sutskever, Kai Chen, Greg Corrado, and Jeffrey Dean.
\newblock 2013.
\newblock Distributed representations of words and phrases and their
  compositionality.
\newblock {\em CoRR}, abs/1310.4546.

\bibitem[\protect\citename{Nallapati \bgroup et al.\egroup }2016]{nallapati}
Ramesh Nallapati, Bing Xiang, and Bowen Zhou.
\newblock 2016.
\newblock Sequence-to-sequence rnns for text summarization.
\newblock {\em ICLR workshop}, abs/1602.06023.

\bibitem[\protect\citename{Neto \bgroup et al.\egroup }2002]{neto:2002:ATS}
Joel~Larocca Neto, Alex~Alves Freitas, and Celso A.~A. Kaestner.
\newblock 2002.
\newblock Automatic text summarization using a machine learning approach.
\newblock In {\em Proceedings of the 16th Brazilian Symposium on Artificial
  Intelligence: Advances in Artificial Intelligence}, pages 205--215.

\bibitem[\protect\citename{Ricardo~Ribeiro}2013]{key_passages}
David Martins de Matos João P. Neto Anatole Gershman Jaime~Carbonell
  Ricardo~Ribeiro, Luís~Marujo.
\newblock 2013.
\newblock Self reinforcement for important passage retrieval.
\newblock In {\em 36th international ACM SIGIR conference on Research and
  development in information retrieval}, pages 845--848.

\bibitem[\protect\citename{Rush \bgroup et al.\egroup }2015]{namas}
Alexander~M. Rush, Sumit Chopra, and Jason Weston.
\newblock 2015.
\newblock A neural attention model for abstractive sentence summarization.
\newblock {\em CoRR}, abs/1509.00685.

\bibitem[\protect\citename{{Sankaran} \bgroup et al.\egroup }2016]{baskaran}
B.~{Sankaran}, H.~{Mi}, Y.~{Al-Onaizan}, and A.~{Ittycheriah}.
\newblock 2016.
\newblock {Temporal Attention Model for Neural Machine Translation}.
\newblock {\em ArXiv e-prints}, August.

\bibitem[\protect\citename{Venugopalan \bgroup et al.\egroup
  }2015]{video_captioning}
Subhashini Venugopalan, Marcus Rohrbach, Jeff Donahue, Raymond~J. Mooney,
  Trevor Darrell, and Kate Saenko.
\newblock 2015.
\newblock Sequence to sequence - video to text.
\newblock {\em CoRR}, abs/1505.00487.

\bibitem[\protect\citename{{Vinyals} \bgroup et al.\egroup
  }2015]{pointer_networks}
O.~{Vinyals}, M.~{Fortunato}, and N.~{Jaitly}.
\newblock 2015.
\newblock {Pointer Networks}.
\newblock {\em ArXiv e-prints}, June.

\bibitem[\protect\citename{Wong \bgroup et al.\egroup }2008a]{wong:2008:ESU}
Kam-Fai Wong, Mingli Wu, and Wenjie Li.
\newblock 2008a.
\newblock Extractive summarization using supervised and semi-supervised
  learning.
\newblock In {\em Proceedings of the 22Nd International Conference on
  Computational Linguistics - Volume 1}, pages 985--992.

\bibitem[\protect\citename{Wong \bgroup et al.\egroup
  }2008b]{linguistic_extractive}
Kam-Fai Wong, Mingli Wu, and Wenjie Li.
\newblock 2008b.
\newblock Extractive summarization using supervised and semi-supervised
  learning.
\newblock In {\em Proceedings of the 22nd Annual Meeting of the Association for
  Computational Linguistics}, pages 985--992.

\bibitem[\protect\citename{Woodsend \bgroup et al.\egroup }2010]{woodsend}
Kristian Woodsend, Yansong Feng, and Mirella Lapata.
\newblock 2010.
\newblock Title generation with quasi-synchronous grammar.
\newblock In {\em Proceedings of the 2010 Conference on Empirical Methods in
  Natural Language Processing}, EMNLP '10, pages 513--523, Stroudsburg, PA,
  USA. Association for Computational Linguistics.

\bibitem[\protect\citename{Zajic \bgroup et al.\egroup }2004]{topiary}
David Zajic, Bonnie~J. Dorr, and Richard Schwartz.
\newblock 2004.
\newblock Bbn/umd at duc-2004: Topiary.
\newblock In {\em Proceedings of the North American Chapter of the Association
  for Computational Linguistics Workshop on Document Understanding}, pages
  112--119.

\bibitem[\protect\citename{Zeiler}2012]{adadelta}
Matthew~D. Zeiler.
\newblock 2012.
\newblock {ADADELTA:} an adaptive learning rate method.
\newblock {\em CoRR}, abs/1212.5701.

\end{thebibliography}
